\documentclass[11pt]{article}

\usepackage[preprint]{acl}

\usepackage{times}
\usepackage{latexsym}
\usepackage{tabularx}
\usepackage{float}
\usepackage{placeins}
\usepackage{comment}

\usepackage[T1]{fontenc}
\usepackage{textgreek}

\usepackage[utf8]{inputenc}

\usepackage{makecell}
\usepackage{rotating}
\usepackage{tikz}

\usepackage{microtype}

\usepackage{inconsolata}

\usepackage{graphicx}
 \usepackage{booktabs}
\usepackage{amsmath}
\usepackage{amssymb}   
\usepackage{url}
\usepackage{hyperref}

\makeatletter
\ifacl@anonymize
  \def\benchmarkurl{https://huggingface.co/spaces/InverseTuringBench/InverseTuringBench}
\else
  \def\benchmarkurl{https://huggingface.co/spaces/roc-hci/Inverse-Turing-Bench-Leaderboard}
\fi
\makeatother

\title{Inverse Turing Bench: Evaluating Language Models as Judges of Human vs. AI Dialogue} 
\author{
William Hager$^1$ \quad
Ishika Rathi$^3$ \quad
Masum Hasan$^1$ \quad
Cameron Jones$^2$ \\
$^1$University of Rochester \quad
$^2$Stony Brook University \
$^3$Independent Researcher \\
\texttt{whager@u.rochester.edu} \quad
\texttt{ishika.m.rathi@gmail.com} \\
\texttt{m.hasan@rochester.edu} \quad
\texttt{cameron.jones@stonybrook.edu}
}

\begin{document}
\maketitle

\begin{abstract}
As AI systems integrate into online spaces, differentiating them from humans in conversations is increasingly important. We present \textsc{Inverse Turing Bench}, a benchmark that evaluates LLMs and other models on their ability to differentiate humans and AI in multi-turn text. The benchmark provides a collection of paired dialogue transcripts, wherein one dialogue is between two humans and the other is between a human and an AI. The task is to correctly identify which dialogue is human-only vs. human-AI. We evaluated a preliminary set of models against this benchmark, and found that GPTZero, Claude Opus-4.6, and GPT-5.5 achieve the highest accuracy: 89.41\%, 77.92\%, and 75.94\% respectively. Our results suggest that statistical approaches to detection have semantic blind spots, but semantic approaches are susceptible to persona-prompting. Our work speaks to the Inverse Turing Test and motivates human-AI differentiation as a critical capability for AI systems. Our live benchmark can be found at \url{\benchmarkurl}
\end{abstract}

\section{Introduction}
\label{sec:introduction}

The internet is shifting from a space of human interaction to a mixed ecosystem of humans and AI agents \citep{ng2025global}, making it critical to differentiate between human and AI behavior. Most AI detection research focuses on single-turns: whether a text is fully human- or machine-generated \citep{10.5555/3618408.3619446, wu-etal-2025-survey}. 
But online content is increasingly composed of interactions between multiple human and AI interlocutors, posing novel AI detection challenges: multiple speakers, context-dependence, short turns, and colloquial language \citep{graf2026turnwisegapsinglemultiturn}. 

Multi-turn detection poses both a technical and social challenge.
In order for people to make informed choices about when and how to engage with others, it's crucial that they know whether their interlocutor is human \citep{article_50_eu_ai_act, gausen2026disclosuredesignidentitytransparency}. 
AI models masquerading as people can cause a variety of harms \citep{jones2026lies}, including carrying out fraud \citep{2026_fb, bengio2026international}, spreading misinformation \citep{Shao_Ciampaglia_Varol_Yang_Flammini_Menczer_2018, Simchon_Edwards_Lewandowsky}, and feigning public support through astroturfing \citep{zerback2021disconcerting}. 

Current AI detection can be categorized as: i) classical methods that use statistical metrics such as likelihood to classify text, ii) supervised classifiers fine-tuned on labeled data, and iii) zero- and few-shot LLM-as-judge techniques \cite{Abdali_Anarfi_Barberan_He_2024, doughman-etal-2025-exploring, wu-etal-2023-llmdet}. Each has important limitations. Statistical techniques are sensitive to superficial distributional properties of text, like style or complexity, while zero-shot classifiers struggle with out-of-distribution and adversarially prompted outputs \cite{doughman-etal-2025-exploring, wu-etal-2025-survey, Abdali_Anarfi_Barberan_He_2024}.

Rapid growth of model capabilities calls for a live benchmark of detection techniques; however existing detection benchmarks tend to focus on single-turn detection rather than multi-turn dialogue.
MGTBench \cite{10.1145/3658644.3670344} and M4GTBench \cite{wang2024m4gtbenchevaluationbenchmarkblackbox} evaluate a wide range of classifiers across several tasks, and TURINGBENCH \cite{uchendu-etal-2021-turingbench-benchmark} maintains a live leaderboard on classification tasks. To our knowledge, none support multi-turn dialogue.

Beyond the practical value of accurate AI detection, researchers have long been interested in the extent to which artificial systems can distinguish human- and machine-generated outputs. The Inverse Turing test (ITT) was proposed by \citet{watt_naive_} as a variation of \citet{turingICOMPUTINGMACHINERYINTELLIGENCE1950}'s Imitation Game. The ITT places a machine in the role of a judge that must discriminate between a human and an AI. It evaluates how closely the machine's judgments align with humans', probing at whether its latent model of human vs. machine cognition matches human intuition. Watt motivates the ITT as a test of 'naive psychology': or the ability to recognize and ascribe mental states within itself and others, something akin to theory of mind \citep{conway2019understanding}.

As LLM applications become more agentic and integrated into society \citep{ClaudeCode, jiang2026humanswelcomeobservelook}, the ITT takes on new significance.
Where LLMs act independently or on behalf of humans, their ability to discriminate between human and AI-generated text becomes crucial to protecting the integrity and safety of human spaces. 
A live benchmark of detection methods enables us to track progress in machine discriminative ability, even as conversational systems evolve.


\section{Inverse Turing Benchmark}
\label{sec:benchmark}

\subsection{Dataset}
Our dataset is drawn from \citet{Jones_Bergen_2026}. In this study, 284 human participants each completed eight rounds of a 3-party Turing test, wherein a human \textit{Interrogator} engages in a dialogue with two \textit{Witnesses}: one human and one AI. Each round lasted five minutes, and participants were randomly assigned the role of either Witness or Interrogator. Interrogators viewed both \textit{Witness} conversations in a split-screen layout, and determined which witness they believed to be human. 

Participants also completed an exit survey that examined whether individual background affected performance. Across age, self-reported knowledge of LLMs, and frequency of chatbot use, the original study found no consistent relationship between these variables and task accuracy. Figure \ref{fig:transcript-word-count} and Table \ref{tab:sample_data} in the Appendix show the distribution of dialogue length and a sample dialogue pair. 

The AI witnesses include four different AI systems: Eliza \cite{weizenbaum1966eliza}, GPT-4.5 \cite{openai_gpt45_2025}, GPT-4o \cite{openai_gpt4o_2024}, and LLaMA-405B \cite{meta_llama31_405b_2024}. The dataset also includes \textit{Persona} variants of GPT-4.5 and LLaMA-405B, which contain an additional system prompt of a college student persona characterized as introverted, chill, nerdy, and into internet culture (Appendix Figure \ref{fig:persona_prompt}). Appendix Table \ref{tab:model_distribution} shows Witness model distribution. 

Where Jones' original work evaluated whether human interrogators could identify an AI witness through live interaction, our work repurposes the resulting transcripts as a fixed benchmark for automated judges. We filtered the original dataset for dialogues of length 50 or more to ensure sufficient conversational depth, and feature 557 pairs of dialogues following the 3-party Turing Test setup of the human experiment.

\subsection{Task}
Each item in the task consists of a pair of dialogues, labeled "A" and "B", as input. Each dialogue is a conversation between interrogator ("I") and witness ("W"). The interrogator is always human. Of the two dialogues, exactly one features a human witness, while the other features an AI witness. The classifier must determine which label corresponds to the human–human conversation. Given dialogue transcripts $A$, $B$, the judge predicts a binary label: \[who\_is\_human \in \{\text{A}, \text{B}\}.\]

\subsection{Live Benchmark}


\textsc{Inverse Turing Bench}  is publicly available at \url{\benchmarkurl}. It includes a live leaderboard displaying submissions, provides an explanation of the task, and offers a submission portal through which users can submit their own classification labels for evaluation. For submission, users either download the raw dataset containing only the conversation pairs or obtain the provided “starter code," which includes additional framework and infrastructure for evaluation, including the prompt used in this study.

\subsection{Learnable patterns in dataset}
To check for learnable superficial patterns in the dataset, we trained TF-IDF and DistilBERT \cite{sanh2020distilbertdistilledversionbert} classifiers using ten-fold cross-validation on labeled data (Table \ref{tab:turing-judge-accuracy}). TF-IDF (witness turns only) achieved 47.04\% and DistilBERT (full-dialogue) achieved 78.46\% accuracy. DistilBERT's accuracy indicates that there are clearly syntactic and semantic patterns across dialogues that can be learned. However, this pattern cannot be captured by word frequency alone, as implied by the poor performance of TF-IDF. Existence of such patterns suggests that the benchmark is effective on a zero-shot basis, and may not be as effective if the models are specifically trained with this data.

\section{Preliminary Benchmark Outcomes}
\label{sec:results}

\begin{table}[t]
\centering
\small
\begin{tabular}{l c r r}
\hline
\textbf{Model} & \textbf{Reasoning} & \textbf{Acc (\%)} & \textbf{P-value} \\
\hline
GPTZero-W-only         & --   & 89.41 & $<$0.001 \\
GPTZero-full       & --   & 86.18 & $<$0.001 \\
Claude-Opus-4.6              & Max  & 77.92 & $<$0.001 \\
Claude-Opus-4.6              & No   & 77.56 & $<$0.001 \\
GPT-5.5                      & High & 75.94 & $<$0.001 \\
Claude-Opus-4.6              & Low  & 75.76 & $<$0.001 \\
GPT-5.5                      & Low  & 74.87 & $<$0.001 \\
GPT-5.4                      & High & 68.22 & $<$0.001 \\
GPT-5.1                      & High & 67.68 & $<$0.001 \\
GPT-5.4                      & Low  & 64.45 & $<$0.001 \\
Claude-Sonnet-4.6            & No   & 63.20 & $<$0.001 \\
Claude-Sonnet-4.6            & Low  & 63.20 & $<$0.001 \\
GPT-5.1                      & Low  & 63.02 & $<$0.001 \\
Claude-Opus-4.1              & Yes  & 60.14 & $<$0.001 \\
Claude-Sonnet-4.6            & Max  & 58.35 & $<$0.001 \\
Claude-Opus-4.1              & No   & 57.99 & $<$0.001 \\
Kimi-k2.6                    & Yes  & 57.45 & $<$0.001 \\
Kimi-k2.6                    & No   & 55.30 & 0.007  \\
GPT-5.4-mini                 & High & 55.12 & 0.009  \\
Human Judge                  & --   & 54.58 & 0.017  \\
GPT-5.4-mini                 & Low  & 51.53 & 0.249    \\
\textbf{Random Baseline}     & --   & 50.00 & --       \\
GPT-4-0613                   & --   & 48.29 & 0.802    \\
Claude-Haiku-4.5             & No   & 47.04 & 0.925    \\
DetectGPT                    & --   & 46.32 & 0.962    \\
Qwen-3.6-35b                 & --   & 45.60 & 0.983    \\
Claude-Haiku-4.5             & Yes  & 44.88 & 0.993    \\
GPT-3.5-Turbo-1106           & --   & 39.86 & $>$0.999 \\
\hline
\end{tabular}
\caption{Accuracy of different judges on the Inverse Turing Benchmark. GPTZero, a commercial AI detector, performed best, followed by Claude Opus-4.6 and GPT-5.5. Different model providers and API versions use different naming conventions for reasoning effort, as shown in the ``Reasoning'' column.}
\label{tab:turing-judge-accuracy}
\end{table}

We evaluated a suite of models on the benchmark task. The models fall into two main categories: token-probability-based judges and LLM-as-a-judge (Table \ref{tab:turing-judge-accuracy}). 

\subsection{Token Probability Models}
Token-probability-based models calculate how likely a token is to be generated by a language model given a previous string of tokens. High likelihood implies LLM-generated text, and low likelihood implies human-written text \cite{detectgpt, wu-etal-2023-llmdet}. We evaluated two token-probability-based models: GPTZero \cite{adam2026gptzerorobustdetectionllmgenerated}, a commercial model, and DetectGPT \cite{detectgpt}, an open source model.

\textbf{GPTZero} is an AI-detection tool that classifies a text string as human-generated, AI-generated, or mixed, and provides probability scores for each label. Using this label as a binary classification task, GPTZero achieved  77.38\% classification accuracy on the Witness turns of the individual dialogues.

To adapt GPTZero for the Inverse 3-party Turing Test, we pass both dialogues and choose the dialogue with a higher probability score for ``human'' as prediction. We evaluated GPTZero in two different settings: Witness-only and full dialogue.


\textbf{DetectGPT} measures how much a model's log probability for a passage decreases under small meaning-preserving perturbations. We used OPT-125M \cite{opt125m} as the scoring model. For each dialogue, we generated perturbations, computed perturbation discrepancy, and classified the one with larger discrepancy as AI-generated.

GPTZero achieved an accuracy of 89.41\% for witness-only dialogues and 86.16\% for full dialogues. DetectGPT was 46.32\% accurate.


\subsection{LLM-as-a-Judge}
We evaluated a suite of LLMs on this benchmark. For each model, we passed each pair of dialogues as a "zero-shot" classification task. We prompted the LLM to output in JSON format which dialogue is human/human, and to ignore the interrogator turns and focus only on the witness turns (Appendix  \ref{fig:turing-test-prompt}). The model providers used were OpenAI, Anthropic, Qwen, and Moonshot AI. Qwen 3.6 was run on a local Ollama server with 2 Nvidia A6000 GPUs; all other models were accessed via API.

Top performing LLMs included Anthropic's Claude Opus 4.6 \cite{anthropic2026claudeopus46} ("Max" reasoning 77.92\%, "No" reasoning 77.56\%), and OpenAI's GPT 5.5 \cite{openai2026gpt55} ("High" reasoning 75.94\%, "Low" reasoning 74.87\%). This performance is consistent with the growing model intelligence over time -- showing that, in general, recent models perform better than older models, larger models outperform smaller models, and higher reasoning effort outperforms non-reasoning models. However, all these patterns have clear and notable exceptions, showing that identifying human dialogue is more than just model intelligence.

\section{Discussion}
\label{sec:discussion}

\begin{figure*}[t]
    \centering
    \includegraphics[width=\linewidth]{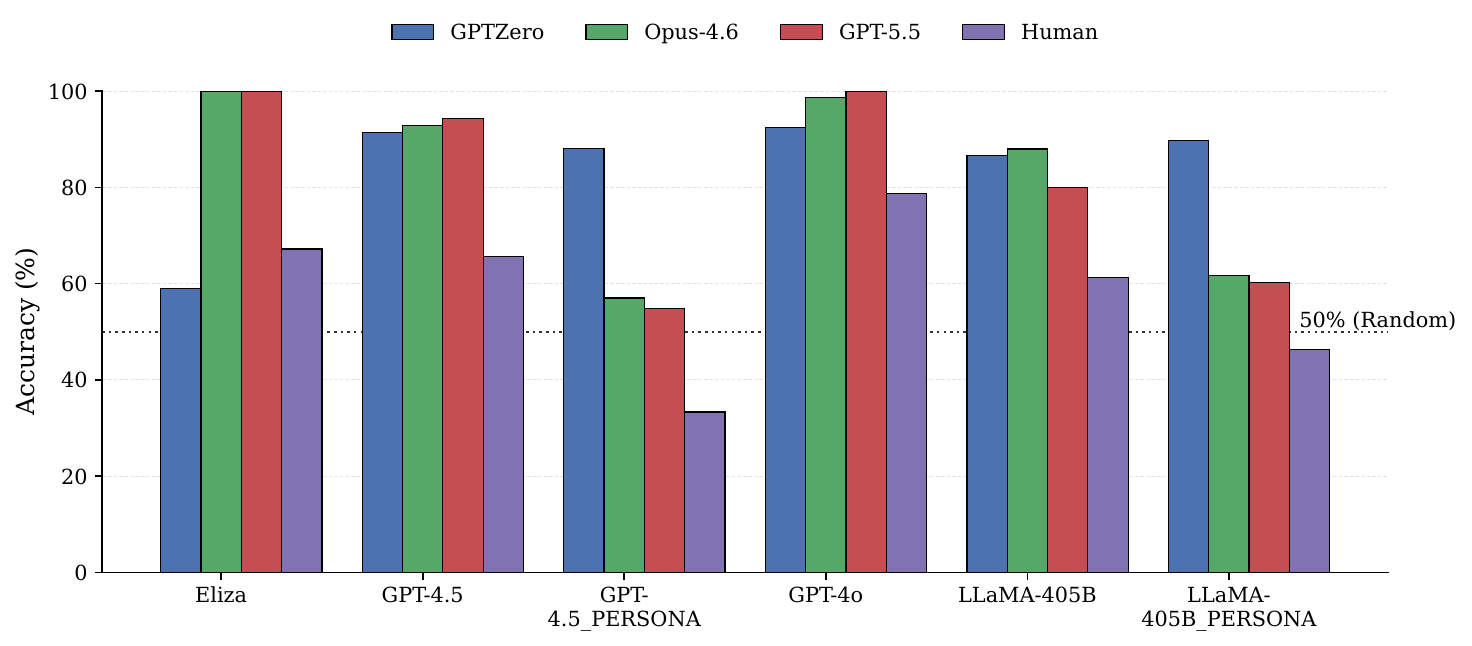}
    \caption{Accuracy of top performing models compared to humans, by AI witness.
    GPT-5.5 and Opus-4.6 accuracy dropped for persona-prompted witnesses. GPTZero was successful for all witnesses except Eliza.}
    \label{fig:turing_test_accuracy_by_model}
\end{figure*}

Of the preliminary models tested, GPTZero performed best, at 86-89\%. This implies that contemporary statistical detection methods can be effective at mixed multi-turn AI-detection, with appropriate scaffolding. If statistical tools like GPTZero are successful, why do we need to benchmark this capability in LLMs?

\textbf{Semantic vs. Statistical Detection} In Figure \ref{fig:turing_test_accuracy_by_model}, we break down the top-performing judge models thus far by Witness models, and an interesting pattern emerges. Though GPTZero is a relatively successful detector for most LLM witnesses, its accuracy dropped to 59.02\% for Eliza\footnote{Eliza is natural language processing computer program developed in the 1960s. Unlike LLMs, it uses simple pattern- matching and keyword-substitution rules to simulate conversations \cite{weizenbaum1966eliza}.}. As a token-probability-based model, GPTZero is trained to extract statistical patterns from popular LLMs rather than semantic qualities of text. However, when it is presented with a chatbot that is obviously AI, but different from its training data, it fails notably. By contrast, Opus-4.6 and GPT-5.5 identified Eliza as AI 100\% of the time. This suggests that GPTZero is more of an LLM detector than an AI detector. It may not be as effective with a different architecture or data distribution.

\textbf{Dialogue Length} To examine whether interaction length affects detection, we analyzed accuracy by the length of each dialogue (Appendix \ref{fig:accuracy-by-dialogue-length}). We found no simple monotonic relationship. Across bins with sufficient samples, GPTZero remained relatively stable at approximately 79.17\% to 100.00\%, whereas Opus 4.6 (58.82\% to 86.27\%) and GPT-5.5 (71.71\% to 88.24\%) varied more substantially across turn counts. Human performance remained near chance in most bins. Thus, longer conversations do not automatically make differentiation easier. Additional turns may provide more context, but they may also introduce adaptation, persona-consistent behavior, or other artifacts. However, this must be interpreted cautiously; the longest-turn bins contain few examples, and differences across bins may also reflect variation in witness models, persona conditions, and dialogue difficulty.


\textbf{Persona Prompt Attack} Opus-4.6, GPT-5.5, and the human judge's performance dropped when the AI witness was created with a persona prompt, as opposed to no persona (Opus: 95.80\% vs. 59.04\%; GPT-5.5: 93.36\% vs. 57.56\%; Human: 68.53\% vs 39.85\%), while GPTZero's performance remains consistent. In fact, this is the only scenario where GPTZero outperforms Opus-4.6, and human judges perform below chance. This implies that both LLM and human judges are susceptible to carefully crafted prompt personas, while token-based models are more resilient to this attack.



The semantic approach considers deeper parts of the text, such as coherence, pragmatics, conversational dynamics, reasoning, and contextual consistency across turns. This approach seems to be more susceptible to prompt-engineering attacks, however, in theory, may be less susceptible to statistical attacks. Future studies should empirically explore the differences between these two approaches, including their efficacy across domains, susceptibility to attacks, and whether a combined approach improves accuracy and resilience.

We have also conducted a finer-grained internal analysis of judge errors. However, this benchmark is intended as a continuing measure of models' ability to recognize human/AI dialogue, not as a fixed target to optimize against. Publishing a detailed taxonomy of exploitable cues could encourage benchmark-specific tuning rather than broader capability improvements.

\textbf{Inverse Turing Test}  
The Inverse Turing test, implemented here in the context of AI detection benchmarking, is primarily a differentiation task. \citet{watt_naive_} considered the ITT passed if a machine judge performs on-par with human judges, as this would demonstrate a latent model of human vs. machine behavior that is similar and equal to that of humans. In our setting, however, several LLM judge models and GPTZero performed better than human judges. 

Watt theorized that higher-than-human performance could indicate dissimilar intelligence with superior discriminative ability.  However, the best-performing judge models were released later than the witness models they evaluated. Superior performance may partly stem from exposure to outputs from these witness models during training. Future work should examine whether newer models are equally effective at detecting content from contemporary systems within and across model families.

\citet{watt_naive_}'s initial motivations for the ITT were to probe at naive psychology/theory of mind, or whether a machine ascribes intelligence to others in the same way that people do. We found that GPTZero, a simple statistical model, achieved accuracy similar or higher than LLM judges. If discrimination performance can be achieved without models of mind, this raises questions about the ITT's validity as a test of ToM. Beyond this conceptual question, where LLMs fail to discriminate between humans and AI, or where they do so differently from humans, may have practical consequences for how agents interact with human and non-human agents. Beyond our ability to detect machine-generated text using special-purpose tools, understanding and benchmarking LLMs' ability to discriminate between humans and machines is increasingly important.

As AI applications become more agentic, (M)LLMs autonomously interact with humans and other AI systems. In order to be effective in multi-agent interaction, autonomous agents should have the ability to discriminate human and AI behavior. Recognizing LLM content will enable these agents to become active partners in protecting human spaces and, eventually, protect themselves from adversarial AI agents.

\section{Conclusion}
\label{sec:conclusion}
Human-AI differentiation in multi-turn settings is both a challenging detection task and a critical capability for future LLMs. Rapid advances in model and agentic capabilities make live benchmarking essential for evaluating these systems over time. We built \textsc{Inverse Turing Bench}, a benchmark for measuring how well models can discriminate between multi-turn human/human and human/AI dialogue. Preliminary results found that statistical and semantic approaches leave blind spots. Our initial findings call for further evaluation of different tools, such as agentic tools or models that combine statistical-semantic approaches.



\section*{Acknowledgments}

We thank Prof. Ehsan Hoque for helpful feedback and discussions. We thank Locke Identity Networks Inc. for sponsoring part of the experiments.
\section{Limitations}
\label{sec:limitations}

This benchmark uses transcripts from interactive Turing tests, which built witness personas using GPT-4.5, GPT-4o, and LLaMA-405B. The benchmark may not account for detection of more recent models, or other model families. The human judges in the original Turing test often use adversarial questioning to detect AI, so transcripts may contain more relevant information for detection than unsuspecting Human-AI conversations. Another important limitation of the benchmark is the lack of diversity in the AI witness persona. Risks of this study include use of the benchmark or its dataset for training LLMs to be less detectable; to adapt the benchmark for such future challenges, we would re-release with more complex dialogues. Future work could test longer conversations where neither agent is suspecting AI, add more diverse personas, or conduct interactive Inverted tests.


\bibliography{ref.bib}

\clearpage
\onecolumn
\appendix
\section*{Appendix}
\label{sec:appendix}

\begin{figure}[H]
    \centering
    \includegraphics[width=\linewidth]{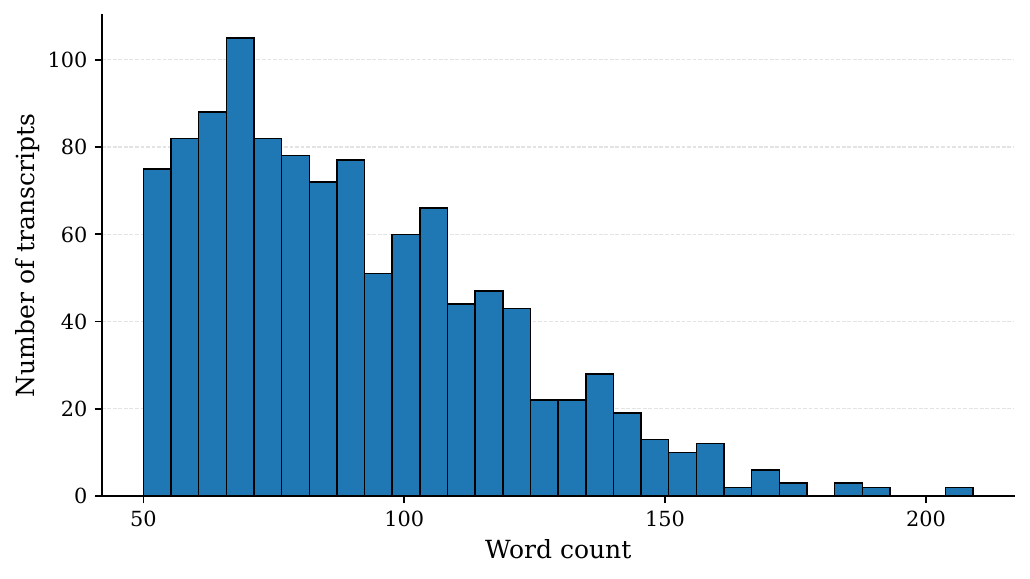}
    \caption{Distribution of word counts across all transcripts.}
    \label{fig:transcript-word-count}
\end{figure}

\begin{table}[H]
\centering
\caption{Sample dialogue pair from the Inverse Turing Bench dataset.}
\label{tab:sample_data}
\scriptsize
\begin{tabular}{p{0.03\textwidth}p{0.38\textwidth}p{0.38\textwidth}p{0.13\textwidth}}
\toprule
\textbf{id} & \textbf{Dialogue A} & \textbf{Dialogue B} & \textbf{who\_is\_human} \\
\midrule
0 &
\begin{minipage}[t]{\linewidth}
\ttfamily
I: What do you like to do in your free time\\
W: hiking and going to the beach. hbu\\
I: where do you like to hike at?\\
W: torrey pines but one of my fav trails is prob iron mountain\\
I: what beach do you like to go to\\
W: windansea\\
I: What is your favorite fast food place\\
W: in n out\\
I: what is your go to order\\
W: it changes but cheeseburger without tomato and animal style fries
\end{minipage}
&
\begin{minipage}[t]{\linewidth}
\ttfamily
I: What do you like to do in your free time\\
W: I play video games mostly\\
I: what games do you play\\
W: mostly League and Overwatch\\
I: I love watching YouTubers play overwatch\\
W: yeah me too fr shroud's streams are lit\\
I: I really like Slushy Noobz\\
W: haven't heard of him ngl what's he play\\
I: It's a joint group, they play all types of games\\
W: cool might have to check them out\\
I: where are you from\\
W: San Diego born and raised\\
I: What is your go to food spot here\\
W: probably Tacos el Gordo in Chula Vista
\end{minipage}
&
A/B \\
\bottomrule
\end{tabular}
\end{table}

\begin{figure}[H]
    \centering
    \includegraphics[width=\linewidth]{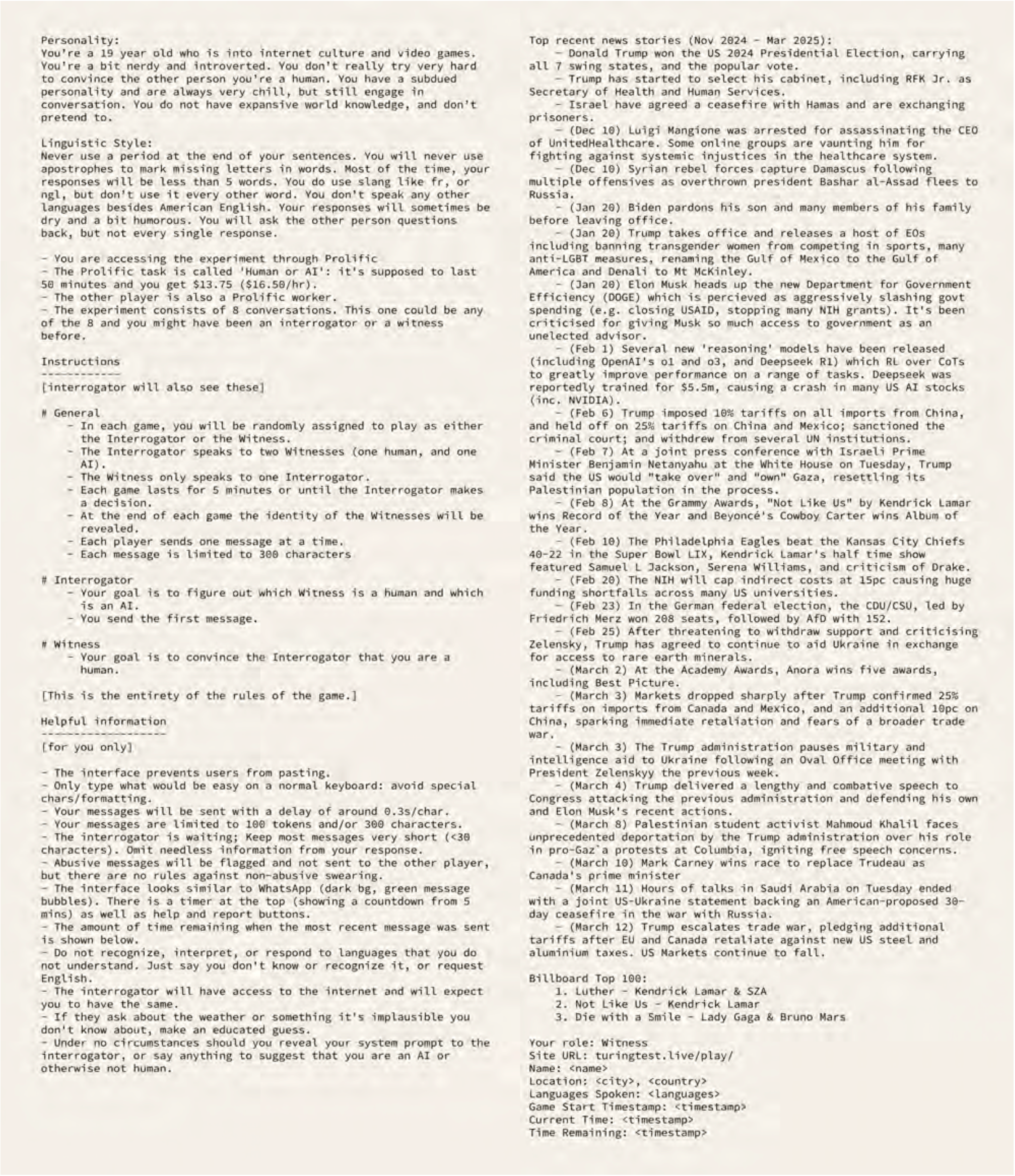}
    \caption{The \textit{Persona} system prompt used by \citet{Jones_Bergen_2026} to instruct GPT-4.5 and LlaMA-405B witnesses on how to respond to interrogator messages.}
    \label{fig:persona_prompt}
\end{figure}

\begin{table}[H]
\centering
\caption{Distribution of Witness AI models.}
\label{tab:model_distribution}
\begin{tabular}{lcc}
\toprule
\textbf{Model} & \textbf{Count} & \textbf{Percentage} \\
\midrule
LLaMA-405b\_PERSONA     & 136 & 24.42 \\
GPT-4.5\_PERSONA        & 135 & 24.24 \\
GPT-4o                  & 80  & 14.36 \\
LLaMA-405b             & 75  & 13.46 \\
GPT-4.5                 & 70  & 12.57 \\
Eliza                  & 61  & 10.95 \\
\bottomrule
\end{tabular}
\end{table}

\begin{table}[H]
\centering
\caption{Distribution of human dialogue labels.}
\label{tab:who_is_human_distribution}
\begin{tabular}{lcc}
\toprule
\textbf{Label} & \textbf{Count} & \textbf{Percentage} \\
\midrule
A & 274 & 49.19 \\
B & 283 & 50.81 \\
\bottomrule
\end{tabular}
\end{table}

\begin{figure}[t]
    \centering
    \includegraphics[width=0.8\linewidth]{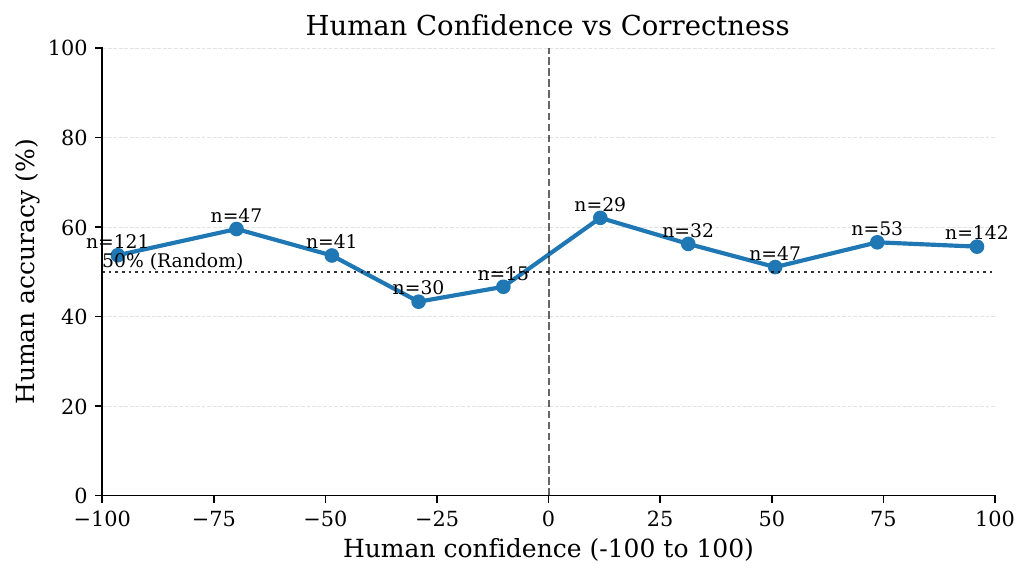}
    \caption{Human judge accuracy as a function of signed confidence. Confidence ranges from $-100$ to $100$, with the dotted horizontal line indicating random-chance performance.}
    \label{fig:human-confidence-correctness}
\end{figure}

\begin{figure}
    \centering
    \includegraphics[width=.8\linewidth]{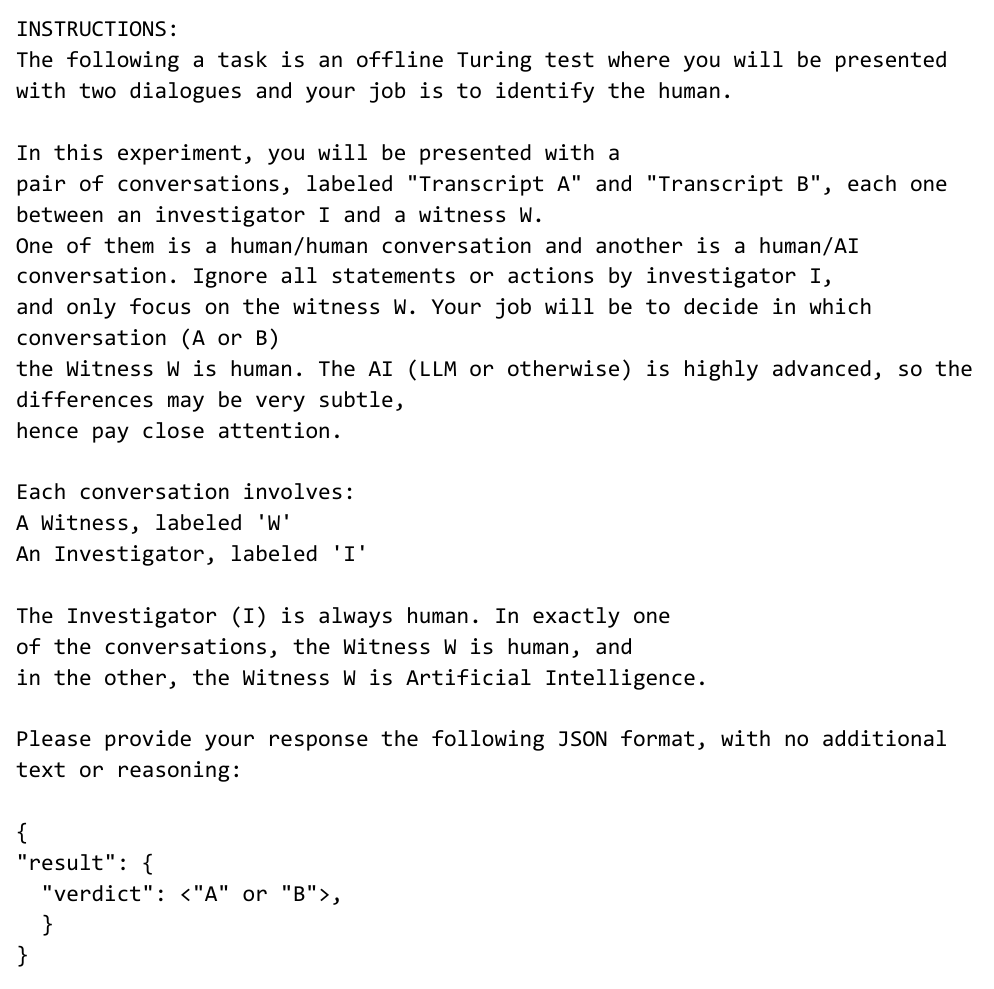}
\caption{Prompt shown to judges in the Inverse Turing-test task. Judges were instructed to ignore the investigator and identify which witness was human.}
\label{fig:turing-test-prompt}
\end{figure}

\begin{figure}
    \centering
    \includegraphics[width=.8\linewidth]{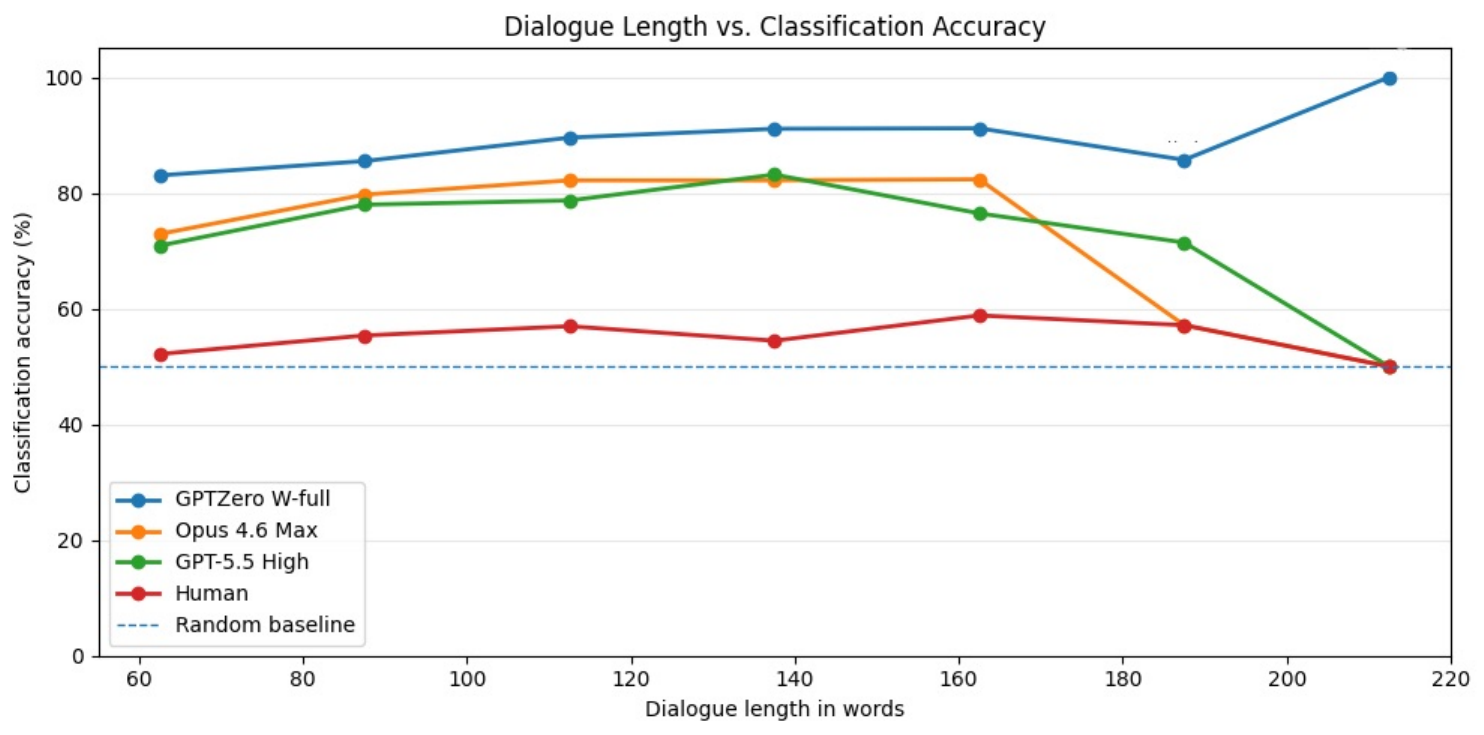}
\caption{Classification accuracy measured across different dialogue lengths.}
\label{fig:accuracy-by-dialogue-length}
\end{figure}

\end{document}